\documentclass[letterpaper]{article}
\usepackage{uai2019}
\usepackage[margin
=1in]{geometry}
\usepackage{moresize}
\usepackage[utf8]{inputenc} % allow utf-8 input
\usepackage[T1]{fontenc}    % use 8-bit T1 fonts
\usepackage{hyperref}       % hyperlinks
\usepackage{url}            % simple URL typesetting
\usepackage{booktabs}       % professional-quality tables
\usepackage{amsfonts}       % blackboard math symbols
\usepackage{nicefrac}       % compact symbols for 1/2, etc.
\usepackage{microtype}      % microtypography
\usepackage{lipsum} 
\usepackage[usenames, dvipsnames]{color}
\usepackage{graphicx}
\usepackage{subcaption}
\usepackage{xspace}
\usepackage{booktabs} % for professional tables
\usepackage{adjustbox}
\usepackage{amsmath}
\usepackage{amssymb}
\usepackage{multirow}
\usepackage{multicol}
\usepackage{algorithm}
\usepackage{algpseudocode}
\usepackage[colorinlistoftodos]{todonotes}
\usepackage{flushend}

\newcommand{\model}{DeepMoE\xspace}
\newcommand{\modelplural}{DeepMoEs\xspace}

\def\vx{{\bf x}}

\renewcommand{\vec}[1]{\mathbf{#1}}

\newcommand{\etal}{{\em et al.}}

\title{Deep Mixture of Experts via Shallow Embedding}

\author{
  Xin Wang$^1$\quad Fisher Yu$^1$\quad 
  Lisa Dunlap$^1$\quad
   Yi-An Ma$^1$\quad 
   Ruth Wang$^1$\quad
  Azalia Mirhoseini$^2$ \\
  \textbf{Trevor Darrell$^1$\quad Joseph E. Gonzalez$^1$} \\
  $^1$EECS Department, UC Berkeley \quad\quad $^2$ Google Brain
}

\begin{document}
\maketitle

\begin{abstract}
Larger networks generally have greater representational power at the cost of increased computational complexity.
Sparsifying such networks has been an active area of research but has been generally limited to static regularization or dynamic approaches using reinforcement learning.  We explore a mixture of experts (MoE)
approach to deep dynamic routing, 
which activates 
certain experts in the network on a per-example basis.  
Our novel \model architecture
increases the representational power
of standard convolutional networks by adaptively sparsifying and recalibrating 
channel-wise features in each convolutional layer. 
We employ a multi-headed sparse gating network to determine the selection and scaling of channels for each input, leveraging exponential combinations of experts within a single 
convolutional network. Our proposed architecture is evaluated on four benchmark
datasets and tasks, and we show that \modelplural are able to achieve higher accuracy with lower
computation than standard convolutional networks.
\end{abstract}

\section{INTRODUCTION}

Increasing network depth has been a dominant trend~\cite{he2016deep} in the design of deep neural networks for computer vision.  
However, increased network depth comes at the expense of computational overhead and increased training time.
To reduce the computational cost of machine translation models, Shazeer et al.~\cite{shazeer2017outrageously} recently explored the design of ``outrageously'' wide sparsely-gated mixture of experts models, which employs a combination of simple networks, called experts, to determine the output of the overall network. 
They demonstrated that these relatively shallow models can reduce computational costs and improve prediction accuracy.
However, their resulting models needed to be many times larger than existing translation models to recover state-of-the-art translation accuracy.
Expanding on this, preliminary work by Eigen et al.~\cite{eigen2013learning} demonstrated the advantages of stacking \emph{two} layers of mixture of experts models for MNIST digits classification.  
With these results, a natural question arises: \emph{can we stack and train many layers of mixture of experts models to improve accuracy and reduce prediction cost without radically increasing the network width?}

In this paper, we explore the design of \emph{deep mixture of experts models (\modelplural)} that compose hundreds of mixture of experts layers.  
\modelplural combines the improved accuracy of deep models with the computational efficiency of sparsely-gated mixture of expert models. 
% In addition, the composition of multiple sparsely-gated expert selection layers yields a potentially exponential increase in the representation capacity with only a linear increase in the number of parameters. 
However, constructing and training \modelplural has several key challenges.
First, mixture decisions interact across layers in the network requiring joint reasoning and optimization.
Second, the discrete expert selection process is non-differentiable, complicating gradient-based training. Finally, the composition of multiple mixture of experts models increases the chance of degenerate (i.e., singular) combinations of experts at each layer.

\begin{figure*}[t]
    \includegraphics[width=\textwidth]{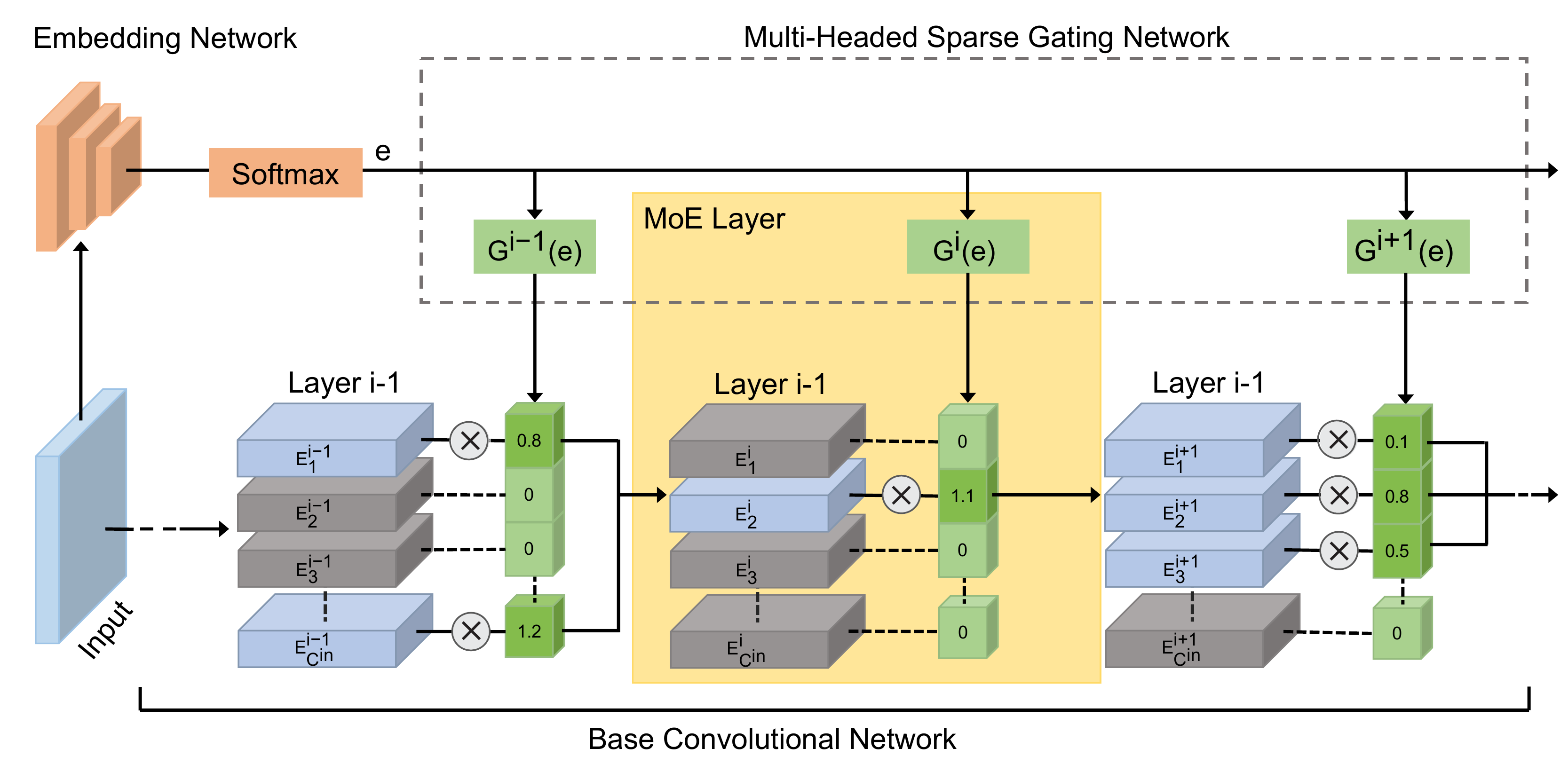}
    \caption{\model architecture. The input is fed into both the base convolutional network and the shallow embedding network. The embedding network outputs the latent mixture of weights, which is then inputted into the multi-headed sparse gating network, to select the experts to activate for that specific layer. The architecture within a single layer of \model strongly resembles the traditional Mixture of Experts structure.}
    \label{fig:moe2}
\end{figure*}

To address these challenges we propose a general \model architecture that combines a deep convolutional network with a shallow embedding network and a multi-headed sparse gating network (see Fig.~\ref{fig:moe2}). The shallow embedding network  terminates in a soft-max output layer that computes \emph{latent mixture weights} over a fixed set of latent experts.
These latent mixture weights are then fed into the multi-headed sparse gating networks (with ReLU outputs)
to \emph{select} and \emph{re-weight} the channels in each layer of the base convolutional network. 
We then jointly train the base model, shallow embedding network, and multi-headed gating network with an auxiliary classification loss function over the shallow embedding network and sparse regularization on the gating network outputs to encourage diversity in the latent mixture weights and sparsity in the layer selection. This helps balance expert utilization and keep computation costs low. 

Recent work~\cite{cohen2016expressive} proves that the expressive power
of a deep neural network increases super-exponentially with its depth, based on the width. 
By stacking multiple mixture of expert layers and dynamically generating
the sparse channel weights, we analyze in Sec.~\ref{sec:expressivity} that
\modelplural reserve the expressive power of the unsparsified deep networks. 
% as long
% as approximately half of the channels in the convolution layers are selected for
% one forward pass. 
% By stacking multiple mixture of experts layers and dynamically generating the channel weights, \model is able to achieve a substantial increase in representational capacity over a single wide mixture of experts layer.
% To characterize this increase in representational capacity, we analyze the combinatorial path structure and show that the expressive power is exponential in the depth of the network (see Sec.~\ref{sec:expressivity}).

Based on this theoretical analysis, we further propose two variants \emph{wide-\model} and \emph{narrow-\model} to improve prediction accuracy
while reducing the computational cost compared to standard convolutional networks.
For wide-\modelplural, we first double the number of channels in the standard
convolutional networks and then replace the widened convolutional layers with
MoE layers. We examine in experiments that if only half of channels in the 
widened layers are selected at inference, wide-\model is able to achieve 
higher prediction accuracy due to the increase of model capacity while maintaining
the same computational cost as the unwidened network. For narrow-\modelplural, 
we directly replace the convolutional layers with MoE layers in the standard
convolution networks which generalizes the existing work~\cite{lin2017runtime} 
on dynamic channel pruning and produces models that are more accurate and 
more efficient than the existing channel pruning literature. 

% \model can be configured to increase accuracy and reduce computation costs.
% When used with wider convolutional networks (i.e., \emph{wide-\model}), \model improves prediction accuracy without increasing the computation cost.
% Alternatively, combined with more typical narrow convolutional networks (i.e., \emph{narrow-\model}), \model substantially reduces prediction costs without significantly impacting accuracy.
% In the later setting, \model generalizes existing work~\cite{lin2017runtime} on dynamic channel pruning and produces models that are both more accurate and more efficient.
 
We empirically evaluate the \model architecture on both image classification and semantic segmentation tasks using four benchmark datasets (i.e.,CIFAR-10, CIFAR-100, ImageNet2012, CityScapes) and conduct extensive ablation
study on the gating behavior and the network design in Sec.~\ref{sec:exp}. We 
find that ~\modelplural achieves the goal of improving the prediction
accuracy with reduced computational cost on various benchmarks. 

Our contributions can be summarized as: (1) We first propose a novel \model design
which allows the network to dynamically select and execute part of the network 
at inference. (2) We theoretically analyze that the proposed \model design preserves
the expressive power of a standard convolutional network with reduced computational cost. (3) We further introduce two
\model variants that are more accurate and efficient than the prior methods on
different benchmarks. 

% where \model is able to achieve higher accuracy with lower computation. 
% In particular, wide-\model improves the ResNet accuracy over 1\% on ImageNet on which the latest shallow MoE work~\cite{gross2017hard} failed to show any improvement. 
% In addition, narrow-\model can be both
% more accurate and efficient than a range of dynamic and static channel pruning techniques and even improve the 
% prediction accuracy over the baselines. The \model design can be handily applied to semantic segmentation task with little modification. In addition to our evaluations of \model, we also conduct extensive ablation studies to understand the gating behaviors of \modelplural and investigate the design choices of \modelplural.

\section{RELATED WORK}
\label{sec:related_work}
\textbf{Mixture of experts.} 
Jacobs et al.~\cite{jacobs1991adaptive} introduced the original formulation of mixture of experts (MoE) models. 
In this early work, they describe a learning procedure for systems composed of many separate neural networks each devoted to subsets of the training data.
Later work~\cite{ collobert2002parallel, collobert2003scaling, jordan1994hierarchical} applied the MoE idea to classic machine learning algorithms such as support vector machines. 
More recently, several~\cite{shazeer2017outrageously, gross2017hard, ahmed2016network} have proposed MoE variants for deep learning in language modeling and image recognition domains.
These more recent efforts to combine deep learning and mixtures of experts have focused on mixtures of deep sub-networks rather than stacking many mixture of expert models.
While preliminary work by Eigen et al.~\cite{eigen2013learning} explored stacked MoE models, they only successfully demonstrated networks up to depth two and only evaluated their design on MNIST Digits. 
In contrast, we construct deep models with hundreds of MoE layers based on a shared shallow embedding rather than the layer outputs~\cite{eigen2013learning} which makes \model more suitable to parallel hardware with  batch parallelism as the gate decisions are pre-determined. 
We also address several of the key challenges around the design and training of multi-layer MoE models.  More recently, the mixture of experts design has been applied in different applications, 
e.g., video captioning~\cite{wang2019learning}, multi-task learning~\cite{ma2018modeling}, etc.

\textbf{Conditional computation.} Related to mixture of experts, recent works by Bengio et al.~\cite{bengio2015conditional, bengio2013estimating, cho2014exponentially} explored conditional computation in the context of neural networks
which selectively executes part of the network based on the input. They use reinforcement learning (RL) for the discrete selection decisions which are delicate to train while our 
sparsely-gated \model design can be embedded into standard convolutional networks and optimized with stochastic 
gradient descent.

\textbf{Dynamic channel pruning.} 
To reduce storage and computation overhead, many~\cite{li2016pruning,he2017channel,luo2017thinet} 
have explored channel level pruning which removes entire channels at each layer in the network and thus leads to 
structured sparsity. However, permanently dropping channels limits the network capacity. 
Bridging conditional computation and channel pruning, recent works~\cite{lin2017runtime,wang2018skipnet,wu2018blockdrop} have explored dynamic pruning, which use per-layer gating networks to dynamically drop individual channels or entire layers based on the output of previous layers. Therefore the channels to be dropped are dependent on the input, resulting in a more expressive network than one that applies static pruning techniques.
Like the work on conditional computation~\cite{wang2017idk}, dynamic pruning relies on sample inefficient reinforcement learning techniques to train many convolutional gates. 
In this work, we generalize the earlier work on dynamic channel pruning by introducing a more efficient shared convolutional embedding and simple ReLU based gates to enable sparsification and feature re-calibration and allowing end-to-end training using stochastic gradient descent.

\section{DEEP MIXTURE OF EXPERTS}
\label{sec:standard_moe}

In this section, we first describe the \model formulation and then introduce the detailed
architecture design and loss function formulation. 

\subsection{MIXTURE OF EXPERTS}

The original mixture of experts~\cite{jacobs1991adaptive} formulation combines a set of
experts (classifiers), $E_1, ..., E_C$, using a mixture (gating) function $G$ that returns a distribution over the experts given the input $\vec{x}$: 
\begin{equation}
   y = \sum_{i=1}^C G(\vec{x})_{i} E_i(\vec{x}).
\end{equation}
Here $G(\vec{x})_i$ is the weight assigned to the $i^\text{th}$ expert $E_i$.
Later work~\cite{collobert2003scaling} generalized this mixture of experts formulation to a non-probabilistic setting where the gating function $G$ outputs arbitrary weights for the experts instead of probabilities. 
We adopt this non-probabilistic view since it provides increased flexibility in re-scaling and composing the expert outputs.

\subsection{DEEPMOE FORMULATION}
\label{sec:s-moe}
In this work, we propose the \model architecture which extends the standard single-layer MoE model to multiple layers within a single convolutional network. While traditional MoE frameworks focus on the model level combinations of experts, \model operates within a single model and treats each channel as an expert.
The experts in each MoE layer consist of the output channels of the previous convolution operation.
In this section, we derive the equivalence between gated channels in a convolution layer and the classic mixture of experts formulation. 

A convolution layer with tensor input $\vec{x}$ having spatial resolution $W \times H$ and $C^\text{in}$ input channels, 
and $C^\text{in} \times k \times k \times C^\text{out} $
convolutional kernel $\vec{K}$ of dimension $k \times k$ can be written as:
\begin{equation}
    \vec{z}_{o, s,t} = \sum_{i=1}^{C^\text{in}} \sum_{u=0}^{k-1} \sum_{v=0}^{k-1} \vec{K}_{i,u,v,o} \vec{x}_{i, s+u, t+v},\label{eq:basicconv}
\end{equation}
where $\vec{z}$ is the $C^\text{out} \times W \times H$ output tensor.
To construct an MoE convolutional layer we scale the input channels by the gate values $\vec{g} \in \mathbb{R}^{C^\text{in}}$ for that layer and rearrange terms:
\begin{align}
    \vec{z}_{o, s,t} &= \sum_{i=1}^{C^\text{in}} \sum_{u=0}^{k-1} \sum_{v=0}^{k-1} \vec{g}_i \vec{K}_{i,u,v,o} \vec{x}_{i, s+u, t+v} \\
    &= \sum_{i=1}^{C^\text{in}} \vec{g}_i \left(\sum_{u=0}^{k-1} \sum_{v=0}^{k-1}  \vec{K}_{i,u,v,o} \vec{x}_{i, s+u, t+v}\right), \label{eq:moeconv} 
    % = \sum_{i=1}^{C^\text{in}} \vec{g}_i E_i\left(\vec{x}_{i}\right).
\end{align}
Defining convolution operator $*$, we can eliminate the summations and subscripts in \eqref{eq:moeconv} to obtain:
\begin{equation}
    \vec{z} = \sum_{i=1}^{C^\text{in}} \vec{g}_i \vec{K}_i * \vec{x}_{i}
    = \sum_{i=1}^{C^\text{in}} \vec{g}_i E_i\left(\vec{x}\right).
    \label{eq:moeconv3}
\end{equation}
Thus, we have shown that gating the input channels to a convolutional network is equivalent to constructing a mixture of experts for each output channel.
In the following sections, we describe how the gate values $\vec{g}$ are obtained for each layer and then present how individual mixture of experts layers can be efficiently composed and trained in the \model architecture.

\subsection{DEEPMOE ARCHITECTURE}
\label{sec:architecture}
\model is composed of three components: a base convolutional network, a shallow embedding network, and a multi-headed sparse gating network. 

The \textbf{base convolutional network} is a deep network where each convolution layer is replaced with an MoE convolution layer as described in the previous section. In our experiments we use ResNet~\cite{he2016deep}
and VGG~\cite{simonyan2014very} 
% and DLA~\cite{yu2017deep} 
as the base convolutional networks. 

The \textbf{shallow embedding network} maps the raw input image into a latent mixture weights to be fed into the multi-headed sparse gating network. To reduce the computational overhead of the embedding network, we use a 4-layer (for CIFAR) or 5-layer (for ImageNet) convolutional network with 3-by-3 filters with stride 2 (roughly 2\% of the computation of the base models).

The \textbf{multi-headed sparse gating network} 
transforms the latent mixture weights produced by the shallow embedding network into sparse mixture weights for each layer in the convolutional network.
The gate for layer $l$ is defined as:
\begin{equation}
    G^l(\vec{e}) = \texttt{ReLU}(W_g^l\cdot \vec{e}),
\end{equation}
% \xin{adding $\cdot$ here}
    where $\vec{e}$ is the output of the shared embedding network \textbf{M} and $W_g^l$ are the learned parameters which, using the ReLU operation,
project the latent mixture weights into \emph{sparse} layer specific gates.

We refer to this  gating design as \emph{on demand gating}.  The number of experts chosen at
each level is data-dependent and the expert selection across different layers can be optimized jointly. 
Unlike the ``noisy Top-K'' design in~\cite{shazeer2017outrageously}, it is not necessary to determine the number of experts at each layer and indeed each layer can learn to use a different number of experts.

\subsection{DEEPMOE TRAINING}
\label{sec:loss}
As with standard convolutional neural networks, \model models can be trained end-to-end using gradient based methods.
The overall goals of the \model are threefold: 
(1) achieve high prediction accuracy, (2) lower computation costs, and (3) keep the network highly expressive. Thus, \model must learn a gating policy that selects a diverse, low-cost mixture of experts for each input.  
To this end, given the input $\vec{x}$ and the target $y$,  we define the learning objective as 
\begin{equation}
    \mathcal{J}(\vec{x};y) = \mathcal{L}_b(\vec{x}; y) + \lambda\mathcal{L}_g(\vec{x}) + \mu \mathcal{L}_e(\vec{x}; y),
    \label{equ:objective}
\end{equation}
$\mathcal{L}_b$ is the cross entropy loss for the base convolutional model, which encourages a high prediction accuracy.

The $\mathcal{L}_g$ term defined:
\begin{equation}
    \mathcal{L}_g(\vec{x}) = \sum_{l=1}^{L}||G^l(M(\vec{x}))||_1,
    \label{equ:reg}
\end{equation}
is used to control the computational cost (via the $\lambda$ parameter) by encouraging sparsity in the gating network. 

Finally, we introduce an additional embedding classification loss $\mathcal{L}_e$, which is the cross-entropy classification loss. This encourages the embedding or some transformation of the embedding to be predictive of the class label, preventing the phenomenon of gating networks converging to an imbalanced utilization of experts~\cite{shazeer2017outrageously}. The intutition behind this loss construction is that examples from the same class should have similar embeddings and thus similar subsequent gate decisions, while examples from different classes should have divergent embeddings, which would in turn discourage the network from over-using a certain subset of channels. 

Because the \model loss is differentiable we train all three sub-networks jointly using stochastic gradient descent.  
Once trained, we then set $\lambda$ and $\mu$ to 0 and continue to train a few more epochs to refine the base convolutional network. 
The full training algorithm is described in Procedure~\ref{algo:training}.

\floatname{algorithm}{Procedure}
\renewcommand{\algorithmicrequire}{\textbf{Input:}}
\renewcommand{\algorithmicensure}{\textbf{Output:}}

\begin{algorithm}
\small{
\caption{Training Algorithm for \model}
\begin{algorithmic}[1]
\Repeat
    \State $e \gets \texttt{EmbeddingNetwork}(x)$
    \For {$i$ from 1 to $L$}
        \State $g^l \gets G^l(e)$
    \EndFor
    \State $output \gets \texttt{BaseNetwork}(x, g^1, \ldots, g^L)$
    \State $\mathcal{L}_b \gets \texttt{CrossEntropy}(output, y) + \lambda \sum_{l=1}^{L}||g^l||_1 + \mu \texttt{CrossEntropy}(e, y)$
    \State Optimize $\mathcal{L}_b$ with SGD
\Until {The model has been trained for $n_0$ epochs}

\State Freeze \texttt{EmbeddingNetwork} and $G^l$ for $l = 1, \ldots, L$, $\lambda \gets 0$,  $\mu \gets 0$
\State Repeat the training loop for another $n_1$ epochs

\end{algorithmic}
\label{algo:training}}
\end{algorithm}

\section{EXPRESSIVE POWER}
\label{sec:expressivity}
The expressive power of deep neural networks is associated with both the width and the depth
of the network. Intuitively, the wider the network is, the more expressive power the network has. Cohen~\etal~\cite{cohen2016expressive} proves that the expressive power of 
a deep neural network increases super-exponentially with respect to the network depth, based on the network width. In 
this section, we demonstrate that due to the dynamic execution nature and the multi-layer stacking
design, \model preserves the expressive power of a standard unsparsified neural network with
reduced runtime computational cost.

% One way to increase the expressive power of a convolutional neural network is to increase the number of channels in the convolutional layers, but the computational cost also increases as the network becomes \emph{wider}. Since channel-pruning is commonly used to reduce computational costs, it is difficult to justify widening a network, as we will have to reduce the expressive power of a network to keep computation costs low.
% To address the trade-off between expressive power and computational cost, we show that the expressive power of \model increases exponentially with the number of layers.
% We outline the theoretical analysis in this section and provide details in the appendix (Sec.A.1).
We define the expressive power of a convolutional neural network as the ability to construct labeling to differentiate input values.
Following Cohen~\etal~\cite{cohen2016expressive}, we view a neural network as a mapping from a particular example to a cost function ( e.g., negative log probability) over labels. 
The mapping can be represented by a tensor
$\mathcal{A}^y$ operated on the combination
of the representation functions. 

More concretely, the rank of $\mathcal{A}^y$,
which scales as $n^{2^L}$ with measure 1 over the space of all possible network parameters
($n$ is the number of channels of a convolutional layer, a.k.a, network width; and $L$ is the network depth), is a measure of the 
expressive power of a neural network 
as established in~\cite{cohen2016expressive}.
In static channel pruning, if $m$ channels 
are kept, then the expressive power of the pruned network becomes $m^{2^L}$ which is 
a strict subspace of $n^{2^L}$ as $m < n$.  

What makes our \model prevail is that (the sparsity pattern of) our mapping $\mathcal{A}^{y}$ depends on the data. We 
prove in the Appendix A.1 that \model has 
an expressive power of $n^{2^L}$ with probability $1-{{m}\choose{n}}^{-L}$ when stacking multiple MoE layers, 
indicating \model preserves the expressive 
power of the unsparsified network. 

Motivated by the theoretical analysis,  
we propose two variants of \model: \emph{wide-\model} and \emph{narrow-\model}. In the former one,
we first increase the number of channels 
in the convolutional networks to increase the expressive power of the network and then 
replace the widened layers with MoE layers. 
By controlling the number of channels selected at runtime, we can improve the 
prediction accuracy with the same amount of
the computation as the unwidened network. 
This design has the potential to be applied to the 
real-world deployment on the new hardware 
architecture supporting dynamic routing, e.g., TPU, where we can 
place a wide network on it and only execute
part of the network at runtime instead of 
placing a static thin network with the same amount of computation.   
Narrow-\model is closer to the 
dynamic channel pruning setting in~\cite{li2016pruning} and comparable
to the traditional static channel pruning.

% To be able to map different data to different values of the cost function, we combine and concatenate simple convolution units together. We show in Sec.A.1 that the bottleneck of a network's expressive power is the number of possible combinations of these convolution units, and thus increasing the depth of the network results in a doubly exponential increase in the possible combinations of mappings based on network width. We also show that while the sparsity of the gating network could reduce the possible combinations, given sufficient diversity in the sparsity, the probability that the sparsity reduces the effective number of combinations decreases exponentially in the network depth. Therefore, because our embedding classification loss encourages variability among experts, we are able to make up for the increased computation costs with a potentially exponentially more expressive network. 

\section{EXPERIMENTS}
\label{sec:exp}
\begin{figure}[t]
    \centering
    \begin{subfigure}[t]{0.14\textwidth}
        \centering
        \includegraphics[width=\textwidth]{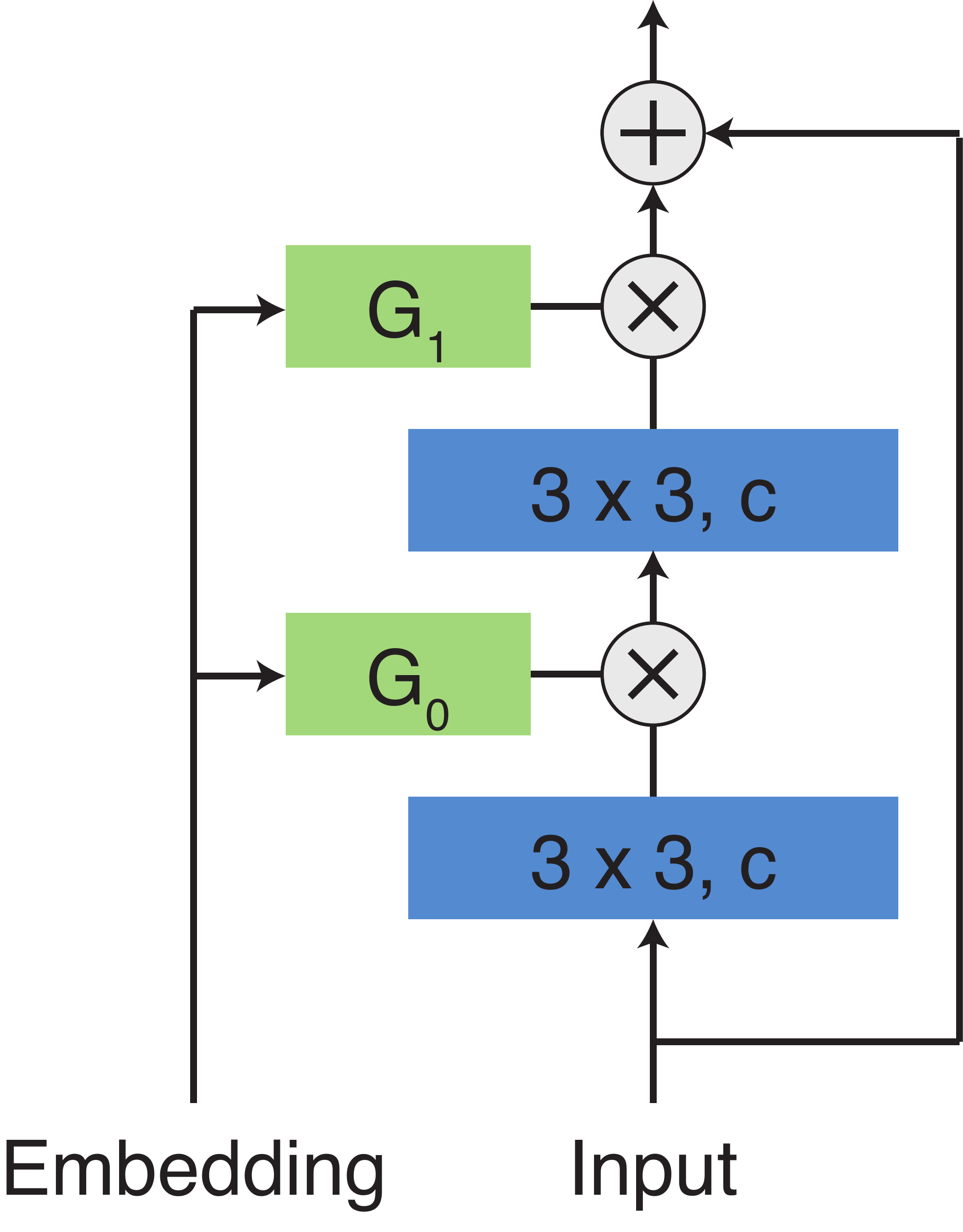}
        \caption{\tiny Basic Block}
        \label{fig:basic_block}
    \end{subfigure}
    ~ 
    \begin{subfigure}[t]{0.14\textwidth}
        \centering
        \includegraphics[width=\textwidth]{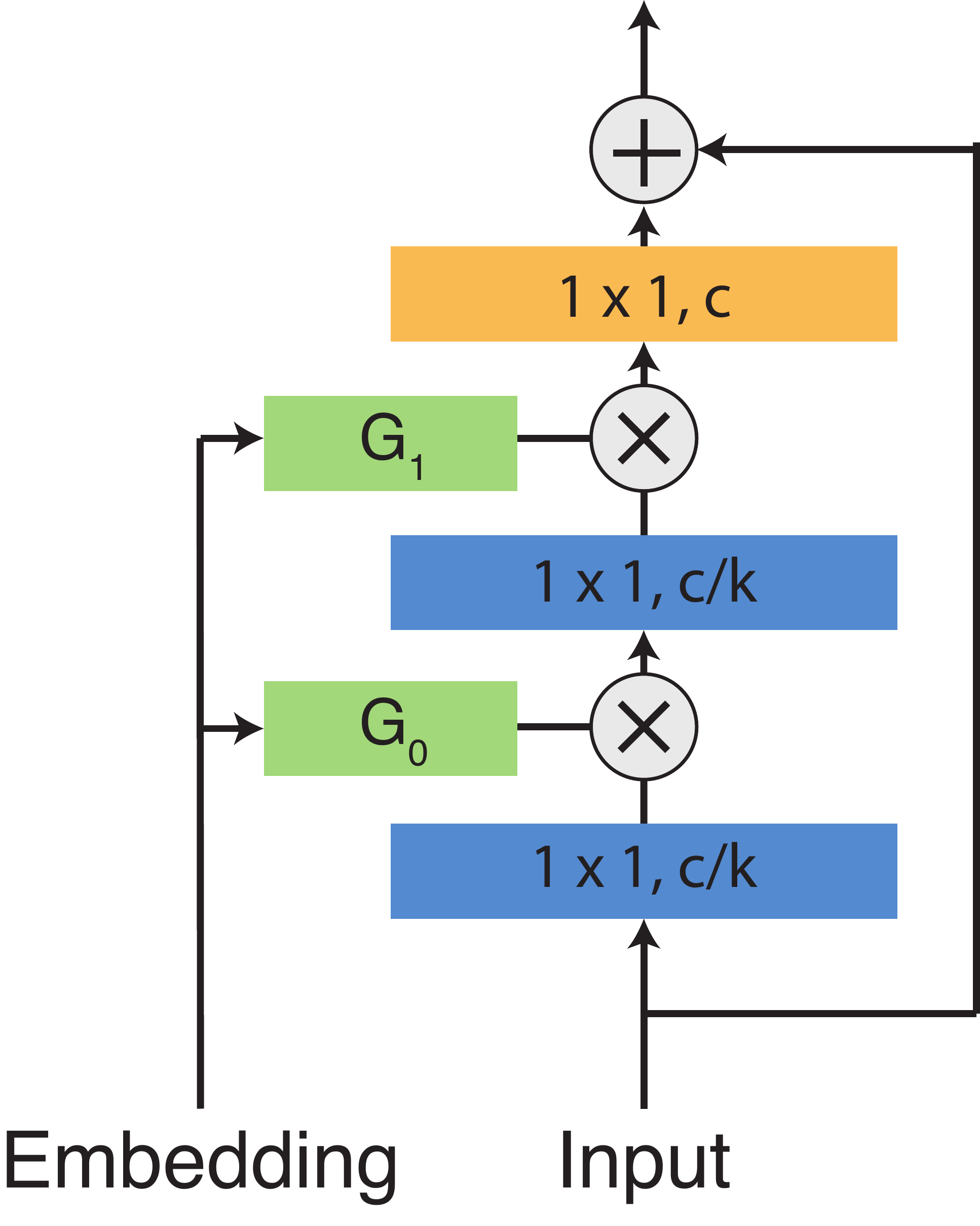}
        \caption{\tiny Bottleneck-A}
        \label{fig:bottleneck_a}
    \end{subfigure}
    ~
    \begin{subfigure}[t]{0.14\textwidth}
        \centering
        \includegraphics[width=\textwidth]{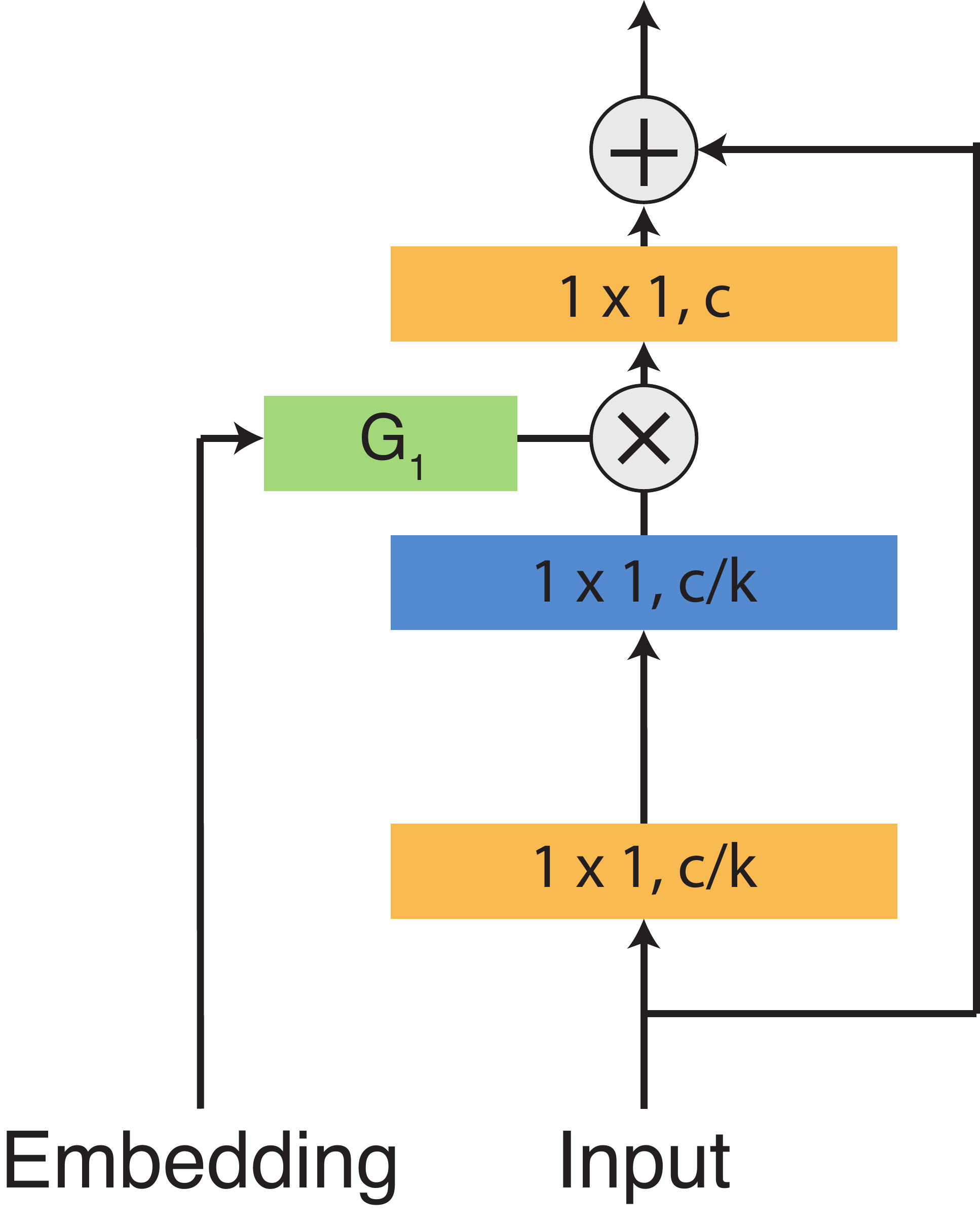}
        \caption{\tiny Bottleneck-B}
        \label{fig:bottleneck_b}
    \end{subfigure}
    \caption{\small Gated Residual Block Designs. The bottleneck-A and B are used in ResNet-50 on
    ImageNet and the basic block for in all the other models. In wide-\model, 
    we increase the number of channels of layers in blue.}
    % (the default is to double the number of channels) }
    \label{fig:block_design}
    % \vspace{-2em}
\end{figure}
In this section, we first evaluate the performance of both wide-\model and narrow-\model on the image classification (Sec.~\ref{sec:wide_moe} and~\ref{sec:narrow_moe}) and semantic segmentation tasks (Sec.~\ref{sec:application}). We observe that \modelplural can achieve lower
prediction error rate with reduced computational cost. 
% Compared to the 
% prior works, our method achieves lower prediction error rate with 
% less computational cost. 
We also analyze the behavior of our gating network, \modelplural regularization effect, and other strategies for widening the network in Sec.~\ref{sec:analysis}.

\textbf{Datasets.} 
For the image classification task, we use the CIFAR-10~\cite{krizhevsky2009learning}, CIFAR-100~\cite{krizhevsky2009learning} and ImageNet 2012 datasets~\cite{russakovsky2015imagenet}. 
For the semantic segmentation task, we use the CityScapes~\cite{Cordts2016Cityscapes} dataset, which provides pixel-level annotations in the images with a resolution of 2048$\times$1024. 
We apply standard data augmentation using basic mirroring and shifting~\cite{wan2013regularization} for CIFAR datasets and scale and aspect ratio augmentation with color perturbation~\cite{xie2017aggregated} for ImageNet. We follow~\cite{yu2017dilated} to enable random cropping and basic mirroring and shifting augmentation for the CityScapes dataset.

\textbf{Models.} 
We examine \model with VGG~\cite{simonyan2014very} and ResNet~\cite{he2016deep} network designs as the base convolutional network (a.k.a, backbone network). 
% We examine \model with a wide range of network designs: VGG network~\cite{simonyan2014very}, ResNet~\cite{he2016deep}. 
VGG is a typical feed-forward network without skip connections and feature aggregation while ResNet, which is composed of many residual blocks, has more complicated connections. To construct \model, we add a gating header after each convolutional layer in VGG and modify the residual blocks in ResNet (Fig.~\ref{fig:block_design}).

In wide-\model, we increase the number of channels in each convolutional layer by a factor of two unless stated otherwise.
In narrow-\model, we retain the same channel configuration as the original base convolutional model. 
% All the models are implemented in PyTorch, and the code will be open-sourced. 

\textbf{Training.} 
To train \model we follow common training practices~\cite{he2016deep, yu2017deep}. 
For the CIFAR datasets, we start training with learning rate 0.1 for ResNet and 0.01
for VGG16, which is reduced by $10\times$ at 150 and 250 epochs with total 350 epochs for the baselines and 270
epochs for \model joint optimization stage and another 80 epochs for fine-tuning with fixed gating networks.

For ImageNet, 
we train the network
with initial learning rate 0.1 for 100 epochs and reduce it by 10$\times$ every 30 epochs. 
We do not further fine-tune the base convolutional network on ImageNet as we find the improvement from fine-tuning is marginal compared to that on CIFAR datasets.

We set the computational cost parameter $\lambda$ in the \model loss function (Eq.~\ref{equ:objective})
between [0.001, 8] (larger values reduce computation) and  $\mu = 1$  
for the CIFAR datasets to match the scale of the cross entropy loss on the base model. For ImageNet we set $\mu=0$ to improve base model feature extraction. 
The training schedule for semantic segmentation is detailed
in Sec.~\ref{sec:application}.

\begin{table}[t]
\small
\centering
\caption{Wide-\model with ResNet-18, ResNet-34, and ResNet-50 on ImageNet. Wide-\model improves the accuracy on ImageNet by $\sim$1\%.}
\label{tab:imagenet}
\adjustbox{width=.72\linewidth}{
\begin{tabular}{c|c}
\toprule
Model &  Top-1 Error Rate (\%) \\
\midrule
ResNet-18 & 30.24 \\
Hard MoE~\cite{gross2017hard} & 30.43\\
Wide-\model-18 & \textbf{29.05}\\
\midrule
ResNet-34 & 26.70 \\
Wide-\model-34 & \textbf{25.87} \\
\midrule
ResNet-50 & 23.85  \\
Wide-\model-50 & \textbf{22.88} \\
\bottomrule
\end{tabular}}
\end{table}

\begin{table}[t]
\small
\centering
\caption{
\small Wide-\model with ResNet-56 and ResNet-100 on CIFAR datasets. Wide-\model improves the prediction accuracy of the baseline ResNet by 3$\sim$4\% on CIFAR-100 and 0.5\% on CIFAR-10.}
\label{tab:cifar}
\adjustbox{width=\linewidth}{
    \begin{tabular}{c|c|c}
    \toprule
    Dataset & Model &  Top-1 Error Rate (\%) \\
    \midrule
    \multirow{2}{*}{CIFAR-10} & ResNet-56 & 6.55 \\
    & Wide-\model-56 & \textbf{6.03}\\
    \midrule
    \multirow{4}{*}{CIFAR-100} & ResNet-56 & 31.46 \\
    & Wide-\model-56 & \textbf{29.77}\\
    \cmidrule{2-3}
    & ResNet-110 & 29.45\\
    & Wide-\model-110 & \textbf{26.14}\\
    \bottomrule
    \end{tabular}
}
\end{table}
% \adjustbox{width=.9\linewidth}{
% \begin{subtable}{.4\textwidth}
%     \caption{CIFAR-100}
%     \label{tab:cifar100}
%     \begin{tabular}{c|c}
%     \toprule
%     Model &  Top-1 Error Rate (\%) \\
%     \midrule
%     ResNet-56 & 31.46 \\
%     Wide-\model-56 & \textbf{29.77}\\
%     \midrule
%     ResNet-110 & 29.45\\
%     Wide-\model-110 & \textbf{26.14}\\
%     \bottomrule
%     \end{tabular}
%         \vspace{-1em}
% \end{subtable}}
% \end{table}

\subsection{WIDE-DEEPMOE}
\label{sec:wide_moe}
In this section, we evaluate the performance of wide-\model as well as its memory usage. 

% \subsubsection{The Effect of Model Capacity on Prediction Accuracy}
\subsubsection{Improved Accuracy with Reduced Computation}
To conduct the evaluation, we first increase
the number of channels in the residual networks by a factor of 2 and then control
the sparsification so that on average half 
of the convolutional channels are selected at 
the inference time. Through our evaluations we find that wide-\model has lower prediction error rate than the standard ResNets on ImageNet (Tab.~\ref{tab:imagenet}), CIFAR-10 and CIFAR-100 (Tab.~\ref{tab:cifar}).

We evaluate ResNet-56 and ResNet-110 on the CIFAR-10 and CIFAR-100
datasets and ResNet-18, ResNet-34, ResNet-50 on ImageNet, using the basic block (Fig.~\ref{fig:basic_block}) for 18 and
34 and bottleneck-A(Fig.~\ref{fig:bottleneck_a}) for 50. The more memory efficient bottleneck-B (Fig.~\ref{fig:bottleneck_b}) is 
also adopted on ImageNet. 

As expressed in Tab.~\ref{tab:imagenet}, wide-\model is able to reduce the error rate of the ImageNet benchmark without
increasing the computational cost (measured by FLOPs) with
networks of different depths. In particular, wide-\model with ResNet-18
and 34 reduce $\sim$ 1\% Top-1 error on ImageNet on which the previous
work~\cite{gross2017hard} fails to show any improvement. 
Similar results can be observed on CIFAR datasets as shown in Tab.~\ref{tab:cifar}, where wide-\model improves the prediction accuracy of the baseline ResNet by 3$\sim$4\% on CIFAR-100 and 0.5\% on CIFAR-10. 

% We observe similar results on CIFAR-100 where wide-\model
% reduces the error of ResNets by 3$\sim$4\%.
% % has a 3$\sim$4\% improvement than the standard ResNet. 
% The improvement on CIFAR-10 is relatively small (0.3\%) compared to other datasets since the 
% opportunity gap between wide-\model and the standard ResNet is small (1.2\%). 

\begin{figure*}[t]
    \centering
        \begin{subfigure}[t]{0.32\textwidth}
        \centering
        \includegraphics[width=\textwidth]{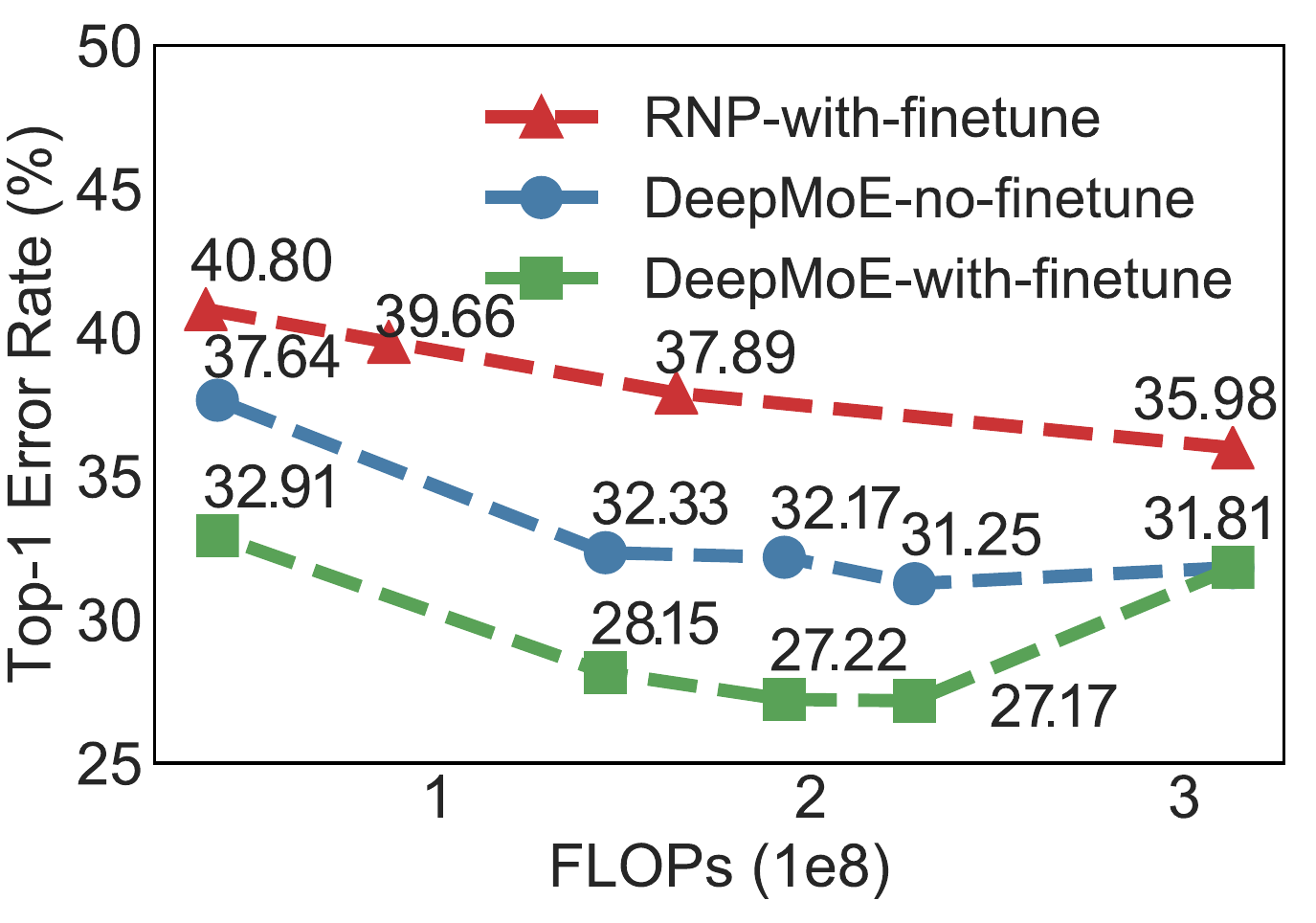}
        \caption{\model \emph{vs} RNP}
        \label{fig:rnp_moe}
    \end{subfigure}
    ~
    \begin{subfigure}[t]{0.32\textwidth}
        \centering   \includegraphics[width=\textwidth]{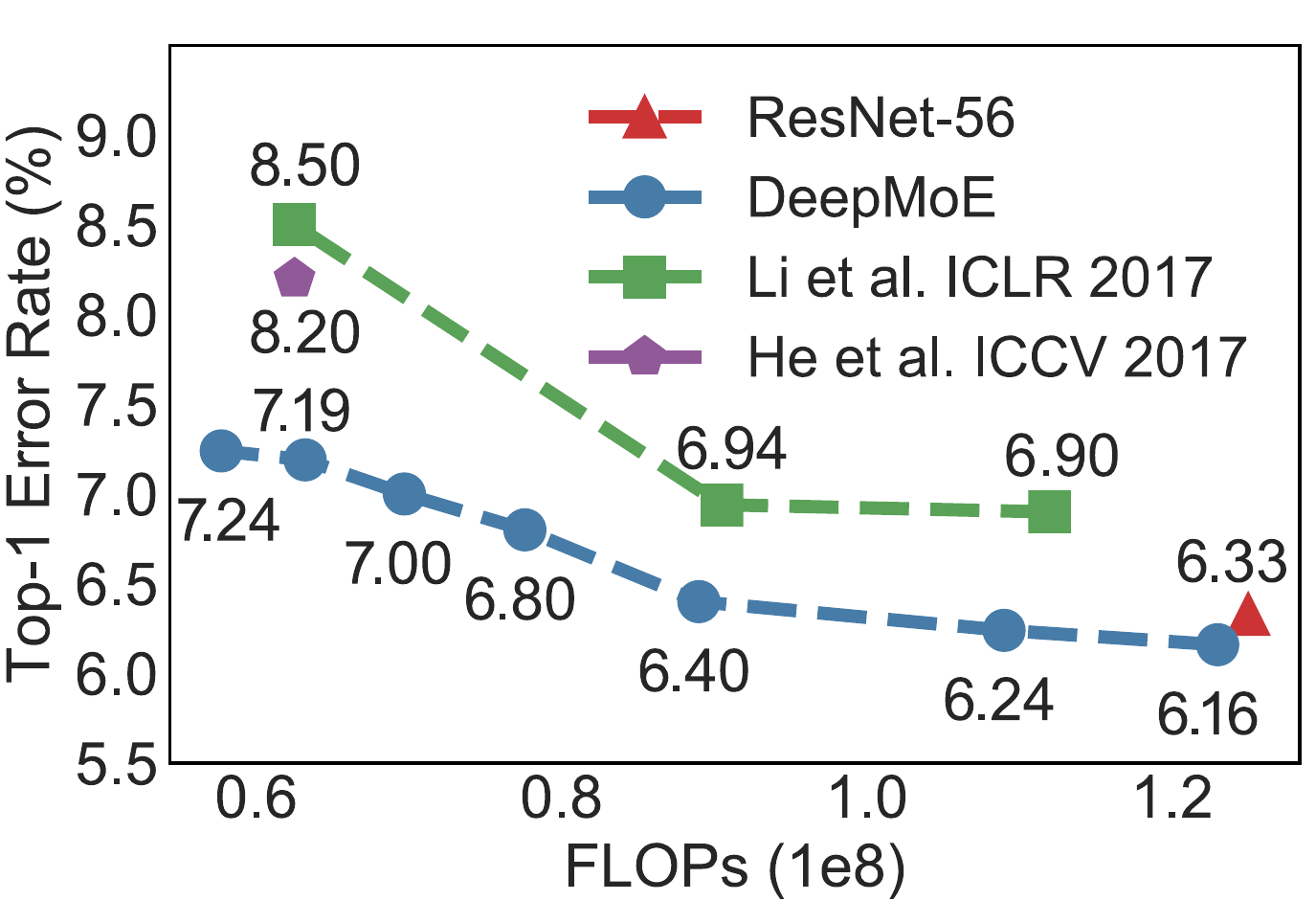}
        \caption{\model \emph{vs} Static Pruning }
        \label{fig:prune_cifar-a}
    \end{subfigure}
    ~ 
    \begin{subfigure}[t]{0.32\textwidth}
        \centering
        \includegraphics[width=\textwidth]{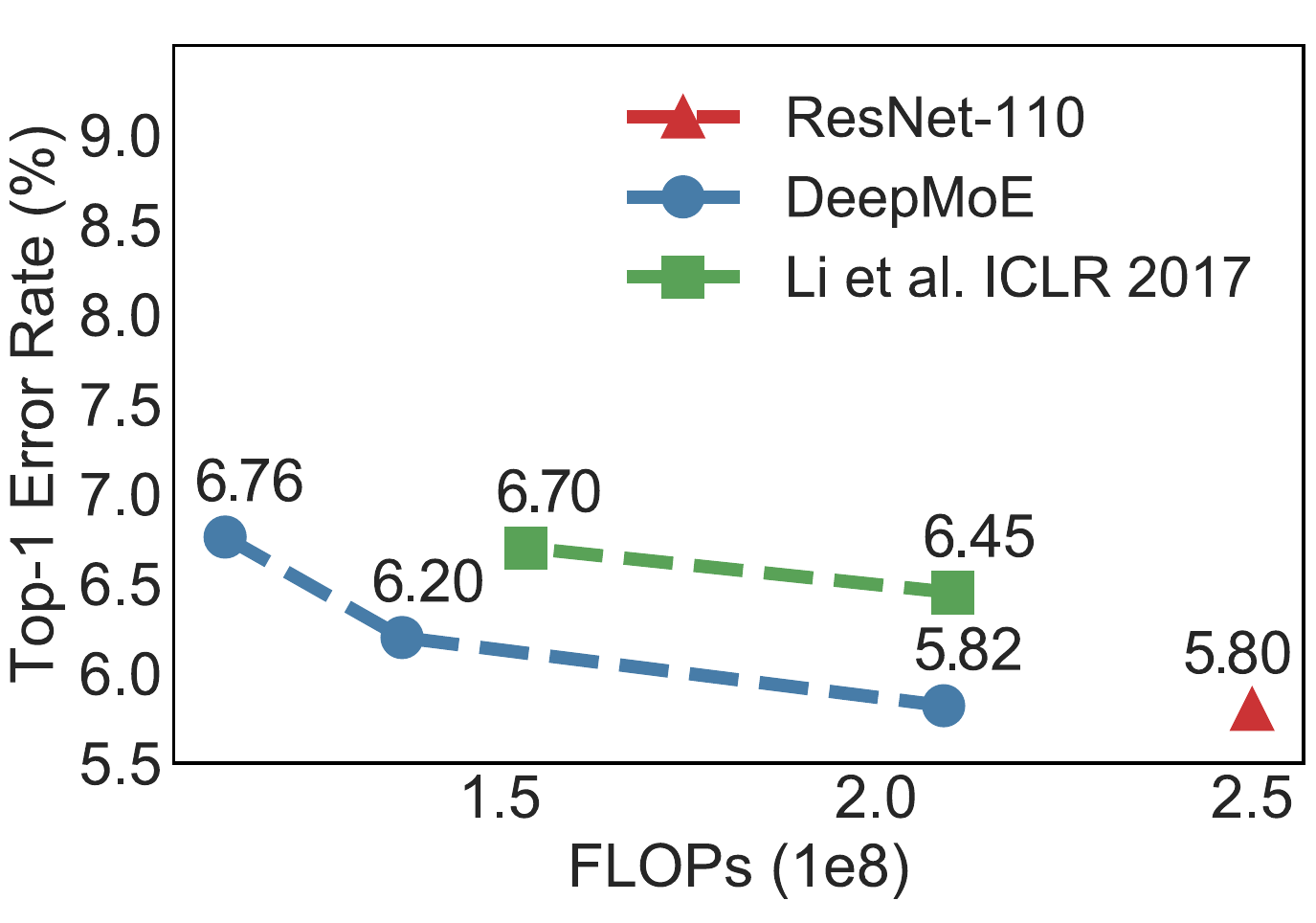}
        \caption{\model \emph{vs} Static Pruning}
        \label{fig:prune_cifar-b}
    \end{subfigure}
    \caption{
    \small (a) \model vs the dynamic pruning approach RNP on CIFAR-100 with VGG16. 
    \model not only outperforms RNP on the accuracy-computation trade-off but improves the accuracy over the baseline VGG model. (b) and (c)  \model vs static pruning approaches on CIFAR-10.}
    \label{fig:cifar10_prune}
\end{figure*}

\subsubsection{Memory Usage}
Another aspect to consider about \model is its memory footprint (proportional to the number of parameters). We examine the memory usage of wide-\model with
widened ResNet-50 as the backbone network and compare it to the standard ResNet-101 which 
has a similar prediction accuracy. We find that wide-DeepMoE using Bottleneck-A (Fig.~\ref{fig:bottleneck_a}) achieves a 22.88\% Top-1 error rate, which compared to ResNet-110 with an error rate of 22.63\%, is only 0.2\% lower in error but requires 20\% less computation. Moreover, wide-\model using Bottleneck-B (Fig.~\ref{fig:bottleneck_b}), which is more memory efficient than Bottleneck-A (Fig.~\ref{fig:bottleneck_a}), achieves 22.84\% top-1 error
with 6\% less parameters and 18\% less FLOPs than the standard ResNet-101 indicating that wide-\model is competitive on the memory usage.

\subsection{NARROW-DEEPMOE}
\label{sec:narrow_moe}
In this section we compare \model to current static and dynamic channel pruning techniques. We show that \model is able to out-preform both dynamic and static channel pruning techniques in prediction accuracy while maintaining or reducing computational costs. 

\subsubsection{Narrow-\model \emph{vs} Dynamic Channel Pruning}
\label{sec:narrow_moe_dynamic}
\model generalizes existing channel pruning work since it both dynamically prunes and \emph{re-scales} channels to reduce the computational cost and improve accuracy. 
In previous dynamic channel pruning work~\cite{lin2017runtime}, channels are pruned based on the outputs of previous layers.
In contrast, the gate decisions in \modelplural are determined in advance based on the shared embedding (latent
mixture weights) which enables improved batch parallelism at inference.

We compare \model to the latest dynamic channel pruning work RNP~\cite{lin2017runtime}  
with VGG-16\footnote{Our baseline 
accuracy is higher than RNP since we use a version with batch normalization in contrast to  the published method.} as the base model on CIFAR-100.
As we can see from Fig.~\ref{fig:rnp_moe}, without fine-tuning, the prediction
error and computation trade-off curve (dotted blue line) of \model is much flatter than RNP (dotted red line) which indicates \model has a greater reduction in  computation without loss of accuracy. 
Moreover, when fine-tuning \model for only 10 epochs (dotted green line in Fig.~\ref{fig:rnp_moe}), \model improves the prediction accuracy by a large margin by 4\% which is a $\sim$ 13\% improvement over the baseline VGG model due to the regularization effect of \model (Sec.~\ref{sec:regularization}).

\subsubsection{Narrow-\model \emph{vs} Static Channel Pruning}
\label{sec:narrow_moe_static}
Similarly, \model outperforms the state-of-the-art static channel pruning results \cite{luo2017thinet,he2016deep,li2016pruning, huang2017data} on both ImageNet shown in Tab.~\ref{tab:resnet-50-imagenet} and the CIFAR-10 dataset in Fig.~\ref{fig:prune_cifar-a} and~\ref{fig:prune_cifar-b}. \model with ResNet-50 reduces
56.8\% of the computation of the standard ResNet-50 with a top-1 error rate of $26.21\%$,  approximately 
 2\% better than He~\etal~\cite{he2017channel}, which currently has  the best accuracy for an equivalent amount of computation among previous work on ImageNet. Fig.~\ref{fig:prune_cifar-a} and~\ref{fig:prune_cifar-b} show that DeepMoE achieves a higher accuracy less computation times than current techniques. 
%  Details are in Tab.~\ref{tab:resnet-50-imagenet} and Fig.~\ref{fig:cifar10_prune}.

\begin{table}[ht]
\small
\centering
\caption{Pruned ResNet-50 on ImageNet. Top-1/5 error rate and
computation FLOPs are reported. DeepMoE is able to achieve a 26.2\% Top-1 error rate, which is 0.6-2.8\% lower than the other models, while using the least amount of computational cost.}
\label{tab:resnet-50-imagenet}
\adjustbox{width=\linewidth}{
\begin{tabular}{l|cccc}
\toprule
Model  & Top-1 & Top-5 &  FLOPs(x$10^9$) & Reduct.(\%) \\
% \midrule
% ResNet-18 & 30.24 & 10.92 & 1.80 & - \\
% \model-DLA-34-v1 & 30.41 & 10.91 & \textbf{0.79} & 74.32\\
\midrule
SSS~\cite{huang2017data} & 26.8 & - & 3.0 & 20.3 \\
Li et al.~\cite{li2016pruning} & 27.0 & 8.9 & 3.0 & 19.0 \\
He et al.~\cite{he2017channel} & - & 9.2 & 1.9 & 50.0 \\
ThiNet~\cite{luo2017thinet} & 29.0 & 10.0 & 1.7 & 55.8 \\
% \model-DLA-34-v2 & 28.22 & 9.42 & 1.69 & 44.78 \\
\midrule
\model & \textbf{26.2} & \textbf{8.4} & \textbf{1.6} & \textbf{56.8} \\
\bottomrule
\end{tabular}}
    \vspace{-1em}
\end{table}

\subsection{ANALYSIS}
\label{sec:analysis}
In this section, we first analyze the effectiveness of the gating behavior in generating embeddings that are predictive of the class label and thus its ability balance expert utilization. We then study the regularization effect of \modelplural sparsification of the channel outputs. Lastly, we explore the effects of widening certain combinations of layers in a network as opposed to widening all convolutional layers as we do in \model. 

\subsubsection{Gating Behavior Analysis}
\label{sec:gate_vis}
To analyze the gating behavior of \model, we evaluate the trained \model with VGG-16 as follows: for a given fine-grained class $A$ (e.g., \emph{dolphin}), we re-assign the
gate embedding for each input in class $A$ with a randomly chosen gate embedding from other classes either within the
same coarse category (referred to as \emph{in-group shuffling}) or different categories (referred to as \emph{out-of-group shuffling}). 

\begin{figure}[t]
    \centering
    \includegraphics[width=.38\textwidth]{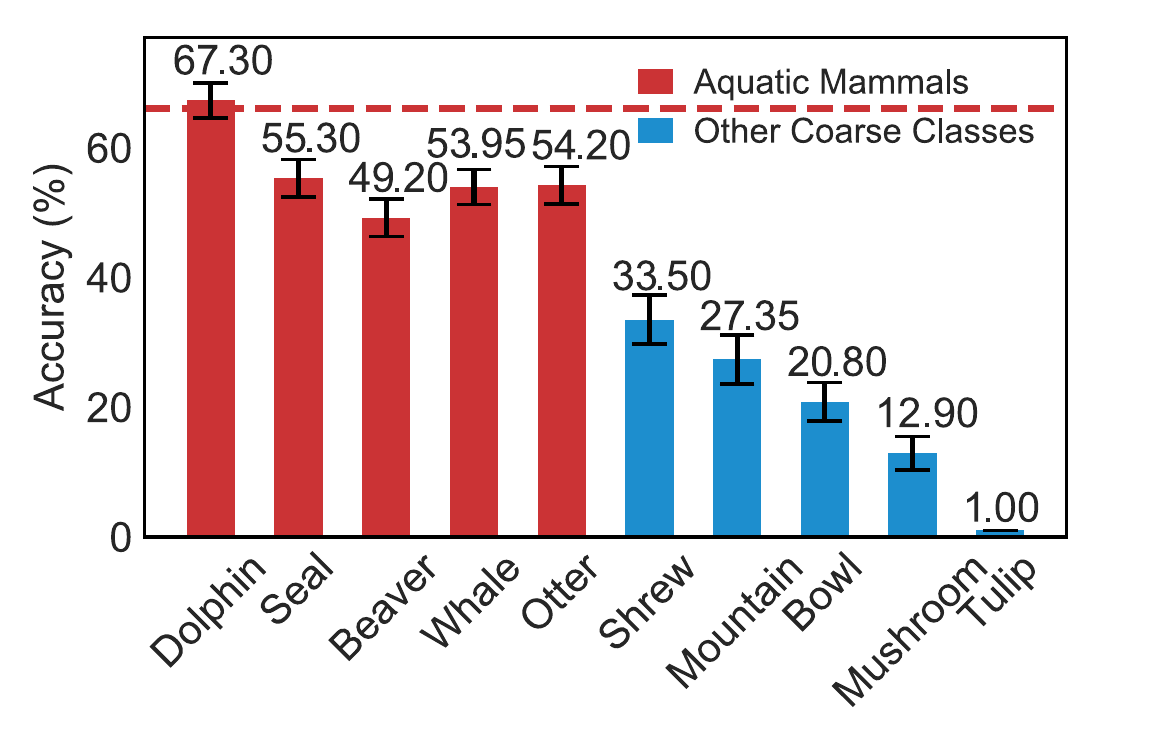}
    \vspace{-1em}
    \caption{\small Gate embedding shuffling. The in-group shuffling has an accuracy 20-65\% higher than out-of-group shuffling.}
     \label{fig:gate_vis}
\end{figure}

In Fig.~\ref{fig:gate_vis},
we plot the test accuracy of class \emph{dolphin}, belonging to the coarse category \emph{aquatic mammals} with randomly selected gate embeddings (repeated 20 times for each input) from 5 classes in the same coarse category (in red) and 5 classes 
for other coarse categories (in blue). 
Fig.~\ref{fig:gate_vis} shows that the test accuracy with in-group embeddings is 20-60\% higher than with out-of-group shuffling.
% The test accuracy with in-group 
% shuffling is much higher than that with out-of-group shuffling.
Especially when applying the gate embeddings from the tulip category,  the test accuracy drops to 1\% while the accuracy with in-group shuffling is 
mostly above 50\%. 
This indicates that the latent mixture of weights are similar for semantically related image categories, and since DeepMoE is never given this coarse class structure, our results are significant.
% This indicates the that the latent mixture weights are similar for semantically related image categories.  
% This result is significant 
% given \model is never given this coarse class structure.

\subsubsection{Regularization Effect of \model}
\label{sec:regularization}
\begin{figure*}[ht]
    \centering
    \includegraphics[width=.85\textwidth]{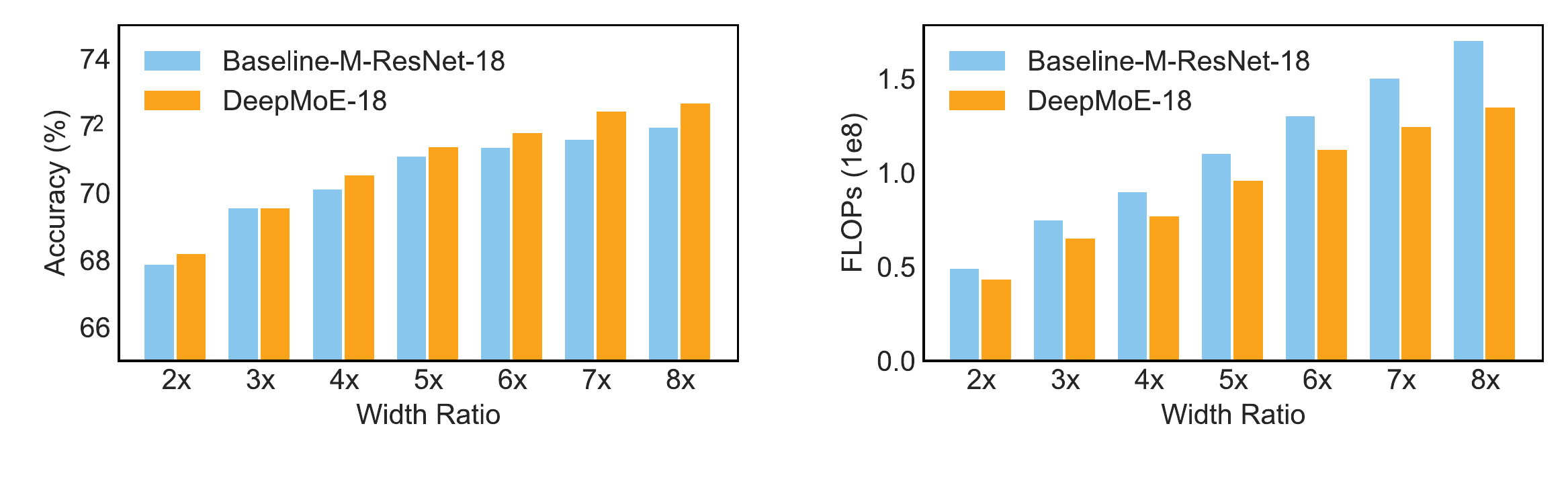}
    \caption{\small Regularization Effect of \model on Widened ResNet. Wide-\model is both more accurate and efficient than the widened baseline models.}
    \label{fig:wide_cifar100}
\end{figure*}

Since \model sparsifies the channel outputs during training and testing, we study the regularization effect of such sparsification. We increase the number of channels of a modified 
ResNet-18 with bottleck-B
(in Fig.~\ref{fig:bottleneck_b}) by 2-8$\times$ on CIFAR-100.
In Fig.~\ref{fig:wide_cifar100}, we plot the accuracy and computation FLOPs of the baseline widened ResNet-18 models (in blue) and wide-\model (in orange) with $\lambda=2$.
% We show the accuracy and the computation FLOPs of the baseline widened ResNet-18 models and wide-\model with $\lambda=2$ in Fig.~\ref{fig:wide_cifar100}.

Fig.~\ref{fig:wide_cifar100} suggests that \model has a lower computation cost and higher accuracy than the baseline widened ResNet, and the advantages of \model increase with the width of the base convolutional network. This indicates a potential regularization effect to the \model design.
% Wide-\model has a lower computation cost and is also more accurate than the baseline wide convolutional network.
% Furthermore, the advantage of \model increases with the width of the base convolutional network suggesting a potential regularizing effect to the \model design.

\subsubsection{\model \emph{vs} Single-Layer MoE}
\label{sec:one_layer_moe}
So far in our experiments, we have widened the network by increasing the number of experts/channels for all convolutional layers. Here, we study other strategies for widening the network. 
We try to widen the VGG-16 model in four different kinds of layers: the top layer (W1-High), the middle layer (W1-Mid),  the lower 4 layers (W4-Low), and finally all the 13 convolutional layers (W13-All) as used in all the other experiments (details in Sec. A.2). 

\begin{table}[t]
    \small
    \centering
    \caption{\small Different widening strategies for VGG16 on CIFAR-100. When controlling the computation FLOPs or both the computation FLOPs and the parameters, the prediction accuracy of widening all convolutional layers is higher than all other widening techniques.}
    \scalebox{0.75}{
    \begin{tabular}{c|c|c|c|c}
    \toprule
     Control & Model & Params  & FLOPs (x$10^8$)  & Acc. (\%) \\
    \midrule
    \multirow{4}{*}{Params} 
    & W1-High & 24.15M & 3.51 & 71.96 \\
    & W1-Mid & 24.16M & 9.18 & 72.02 \\
    & W4-Low  & 24.18M & 43.16 & 72.51 \\
    & W13-All   & 24.18M & 8.15 & \textbf{73.91} \\
    \midrule
    \multirow{4}{*}{Params \& FLOPs} 
    & W1-High & 24.15M & 2.98 & 73.28 \\
    & W1-Mid & 24.16M & 2.74 & 72.68\\
    & W4-Low  & 24.18M & 2.45 & 73.33 \\
    & W13-All   & 24.18M & 2.29 & \textbf{73.39} \\    
    \bottomrule
    \end{tabular}
    }
    \label{tab:one_moe_control}
\end{table} 
As shown in Tab.~\ref{tab:one_moe_control}, the prediction accuracy of W13-All is strictly better than that of a single-layer MoE, even though they have the same number of parameters. Adding MoE to the bottom or top layers is more effective than adding it to the middle layer. Alternatively, if we control both the number of parameters and the computation FLOPs, the accuracy differences between different strategies are reduced but W13-All is still favorable to other widening strategies. 

\newcommand\ver[1]{\small\rotatebox{90}{#1}}
\setlength{\tabcolsep}{3pt}
\begin{table*}[t]
    \small
    \centering
       \caption{\small Segmentation Results on
       CityScapes. The more efficient version {wide-\model-50-a} beats the baseline by 1.5\% of mIoU with a slight increase in FLOPs, while the more accurate version {wide-\model-50-b} outperforms the wide baseline by almost 2\% of mIoU with lower FLOPs.}
       \vspace{1em}
    \scalebox{1.0}{
    \resizebox{\textwidth}{!}{%
\begin{tabular}{@{}ccccccccccccccccccccccccccccccccccccccccc|c|c@{}}
% \toprule
              & \ver{Road} & \ver{Sidewalk} & \ver{Building} & \ver{Wall} & \ver{Fence} & \ver{Pole} & \ver{Light} & \ver{Sign} & \ver{Vegetation} & \ver{Terrain}  & \ver{Sky} & \ver{Person} & \ver{Rider} & \ver{Car} & \ver{Truck} & \ver{Bus} & \ver{Train} & \ver{Motorcycle} & \ver{Bicycle} & {mIoU} & {FLOPs($\times10^9$)}  \\ \midrule
DRN-A-50      & 96.9       & 77.4           & 90.3           & 35.8       & 42.8        & 59.0       & 66.8        & 74.5       & 91.6             & 57.0   & 93.4      & 78.7         & 55.3        & 92.1      & 43.2        & 59.5      & 36.2        & 52.0             & 75.2          & 67.3           & \textbf{703} \\
wide-DeepMoE-50-A    & 97.2       & 78.9           & 90.3           & 45.6       & 48.4        & 56.2       & 61.6        & 72.9       & 91.6             & 60.7 
& 94.2      & 77.4         & 50.6        & 92.5      & 48.7        & 68.7      & 44.1        & 52.7             & 74.2          & \textbf{68.8}           & 804 \\ \midrule
wide-DRN-A-50 & 97.4       & 80.6           & 90.6           & 38.5       & 49.0        & 58.7       & 65.1        & 73.4       & 91.8             & 59.5   & 93.9      & 78.2         & 51.1        & 92.9      & 49.1        & 68.7      & 51.3        & 52.2            & 74.5          & 69.3           & 2173        \\
wide-DeepMoE-50-B    & 97.5       & 80.4           & 91.0           & 48.9       & 50.6        & 58.5       & 65.7        & 75.3       & 92.0             & 60.1  
& 94.7      & 79.2         & 54.7        & 93.2      & 53.8        & 73.2      & 53.2        & 54.8             & 75.6          & \textbf{71.2}           & \textbf{1738} \\ \bottomrule
\end{tabular}
}
}
 \label{tab:seg_cityscapes}
\end{table*}

\subsection{SEMANTIC SEGMENTATION}
\label{sec:application}
Semantic image segmentation requires predictions for each pixel, instead of one label for the whole image in classification. We evaluate \model on the segmentation task to understand its generalizability. In specific, we apply \model to DRN-A~\cite{yu2017dilated}, which adopts ResNet architecture as the backbone, and evaluate the results on the popular segmentation dataset CityScapes~\cite{Cordts2016Cityscapes}. We follow the same training procedure as Yu~\etal~\cite{yu2017dilated} for fair comparison.
The optimizer is SGD with momentum 0.9 and crop size 832. The starting learning rate is set to 5e-4 and divided by 10 after 200 epochs.
The intersection-over-union (IoU) scores and computation costs in FLOPs of \model are presented in Tab.~\ref{tab:seg_cityscapes}.

The hyper-parameter $\lambda$ can adjust the trade-offs between computer efficiency and prediction accuracy. Our efficient model \texttt{wide-\model-50-A} beats the baseline by 1.5\% of mIoU with a slight increase in FLOPs, while our accurate model \texttt{wide-\model-50-B} outperforms the wide baseline by almost 2\% mIoU with lower FLOPs. These results indicate \model is effective on pixel-level prediction such as semantic segmentation as well as image classification. 

\section{CONCLUSION}
In this work we introduced our design of deep mixture of experts models, which produces a more accurate and computationally inexpensive model for computer vision applications.
Our \model architecture leverages a shallow embedding network to construct latent mixture weights, which is then used by sparse multi-headed gating networks to select and re-weight individual channels at each layer in the deep convolutional network. 
This design in conjunction with a novel sparsifying and diversifying loss enabled joint differentiable training, addressing the key limitations of existing mixture of experts approaches in deep learning. We provided theoretical analysis on the expressive power of \model and proposed two 
design variants. The extensive experimental evaluation indicated that \model can reduce computation and surpass accuracy over baseline convolutional networks, as well as improving upon the residual network result on the challenging ImageNet benchmark by a full 1\%. 
Through our analysis we were also able to prove that our embedding and gating network is able to resolve coarse grain class structure in the underlying problem. This work shows promising results when applied to semantic segmentation tasks, and could be incredibly useful for various other problems.

\newpage
% \section*{References}
\bibliography{references}
\bibliographystyle{abbrv}
\newpage
\section*{APPENDIX}

\subsection*{A.1 Expressive Power of \model}
To characterize the expressive power of \model, we follow the tensor analysis approach of Cohen et al.~\cite{cohen2016expressive}.
We first represent an instance of data as a collection of vectors $(\vx_1,\cdots,\vx_N)$, where $\vx_i \in \mathbb{R}^s$.
For the image data, the collection $(\vx_1,\cdots,\vx_N)$ corresponds to vector arrangements of possibly overlapping patches around pixels.
We represent different features in data using (positive) representation functions: 
\begin{align}
f_{d}(\vx_i), \label{f}
\end{align}

so that the convolution operations over data become multiplications over representation functions.
For the representation functions, index $d\in \{1,\cdots,M\}$, where $M$ is the number of different features in data that we wish to distinguish and can be combinatorially large with respect to the number of pixels.

For classification tasks, we view a neural network as a mapping from a particular instance to a cost function (e.g., the log probability) over labels $y$ for that instance.
With the new representation of data instances following Eq.~\eqref{f}, the mapping can be represented by a tensor $\mathcal{A}^y$ operated on the combination of the representation functions:
\begin{align}
    h_y(\vx_1,\cdots,\vx_N) = \sum_{d_1,\cdots,d_N=1}^M \mathcal{A}_{d_1,\cdots,d_N}^y \prod_{i=1}^N f_{d_i}(\vx_i).
\end{align}
To be able to distinguish data instances $\vx$ from $\tilde\vx$, we need $h_y(\vx_1,\cdots,\vx_N) - h_y(\tilde\vx_1,\cdots,\tilde\vx_N)$ to be nonzero. For a fixed mapping $\mathcal{A}^y$, this requirement is equivalent to:
\[
\sum_{d_1,\cdots,d_N=1}^M \mathcal{A}_{d_1,\cdots,d_N}^y \left(\prod_{i=1}^N f_{d_i}(\vx_i) - \prod_{i=1}^N f_{d_i}(\tilde\vx_i) \right) \neq 0,
\]
for $\vx\neq\tilde\vx$.
It can directly be seen that the inequality is satisfied when the difference $\prod_{i=1}^N f_{d_i}(\vx_i) - \prod_{i=1}^N f_{d_i}(\tilde\vx_i)$ is not in the null space of $\mathcal{A}_{d_1,\cdots,d_N}^y$.
Therefore, the expressive power is equivalent to the \emph{rank} of the tensor $\mathcal{A}^y$.
This approach, taken by~\cite{cohen2016expressive}, establishes that for a certain type of networks, the rank of $\mathcal{A}^y$ scales as $n^{2^{L}}$ with measure $1$ over the space of all possible network parameters, where $n$ is the number of channels between network layers (width) and $L$ is the network depth.

If we directly apply the theorem to a wider network (width $m$ satisfying $m>n$), then the rank of $\mathcal{A}^y$ will scale as $m^{2^{L}}$, which is $\left(\dfrac{m}{n}\right)^{2^{L}}$ times better.
However, when the channels are gated with \emph{static} sparse weights, the set of $\mathcal{A}^y$ with this restriction has measure $0$ in the overall space of network parameters. 
In fact, if the number of nonzero weights over the channels is $n$, then the rank of $\mathcal{A}^y$ still scales as $n^{2^{L}}$.
%To prove this, 

What makes our \model prevail is that (the sparsity pattern of) our mapping $\mathcal{A}^y$ depends on the data.
We hereby compare an $L$-layer \model with width equal to $m$ and number of nonzero weights over the channels equal to $n<m$ against an $L$-layer fixed, non-sparse neural network with width equal to $m$.
For the latter, we know that it will be able to distinguish between features in a subspace of dimension $m^{2^{L}}$.
For the former, if $h_y(\vx_1,\cdots,\vx_N) - h_y(\tilde\vx_1,\cdots,\tilde\vx_N) \neq 0$ for the same choices of features (in the $m^{2^{L}}$ dimensional subspace), then we know that it will have expressive power of at least $m^{2^{L}}$:
\begin{align}
&h_y(\vx_1,\cdots,\vx_N) - h_y(\tilde\vx_1,\cdots,\tilde\vx_N) \nonumber\\
=& \sum_{d_1,\cdots,d_N=1}^M \mathcal{A}_{d_1,\cdots,d_N}^y \prod_{i=1}^N f_{d_i}(\vx_i)
-\\
& \sum_{d_1,\cdots,d_N=1}^M \mathcal{\tilde A}_{d_1,\cdots,d_N}^y \prod_{i=1}^N f_{d_i}(\tilde\vx_i) \nonumber\\
=&
\sum_{d_1,\cdots,d_N=1}^M \mathcal{A}_{d_1,\cdots,d_N}^y \left(\prod_{i=1}^N f_{d_i}(\vx_i) - \prod_{i=1}^N f_{d_i}(\tilde\vx_i) \right)  \label{L2}  \\  \label{L3}
&+ \sum_{d_1,\cdots,d_N=1}^M \left(\mathcal{A}_{d_1,\cdots,d_N}^y - \mathcal{\tilde A}_{d_1,\cdots,d_N}^y\right) \prod_{i=1}^N f_{d_i}(\tilde\vx_i).
\end{align}
Since the gating network is independent from the convolution neural network, to have Line~\eqref{L2} exactly equal to the negative of Line~\eqref{L3}---when they are both nonzero---has zero measure over the space of network parameters (even with the sparsity constraint).
We simply need to focus on the cases where Line~\eqref{L2} is zero for the pair of $\vx$ and $\tilde\vx$, and discuss whether Line~\eqref{L3} is also zero.
%
%For an $L$ layer network with width equal to $m$ and number of nonzero weights over the channels equal to $n<m$, rank of $\mathcal{A}_{d_1,\cdots,d_N}^y$ is $n^{2^{L}}$.
In those cases, we assume that the sparsity pattern of the weights over the gated channels is i.i.d. with respect to each channel.
With this assumption, probability of choosing exactly the same channels for different data: $\mathcal{A}_{d_1,\cdots,d_N}^y = \mathcal{\tilde A}_{d_1,\cdots,d_N}^y$ is ${{m}\choose{n}}^{-L}$. 
When they are not equal, the difference $\mathcal{A}_{d_1,\cdots,d_N}^y - \mathcal{\tilde A}_{d_1,\cdots,d_N}^y$ 
can be represented as combinations of linearly independent basis in $\mathbb{R}^{M^N}$ and
positivity of the representation functions ensures that Line~\eqref{L3} is not zero with probability $1$. 
Therefore, $h_y(\vx_1,\cdots,\vx_N) - h_y(\tilde\vx_1,\cdots,\tilde\vx_N) \neq 0$ holds with probability $1-{{m}\choose{n}}^{-L}$.
In other words, there is a $1-{{m}\choose{n}}^{-L}$ probability that the expressive power of our \model equals to or is bigger than $m^{2^{L}}$.

\subsection*{A.2 Network Configurations of Wide VGG}
In Sec.~\ref{sec:one_layer_moe}, we conduct experiments to investigate different widening strategies. We
used four different strategies to widen the VGG-16 network which contains 13 convolutional layers in total: W1-High widens the top layer only, W1-Mid widens the middle layer only, W4-Low widens the lower 4 layers, and finally W13-All that widens all 13 convolutional layers in Tab.~\ref{tab:config}. 

\begin{table}[h]
    \centering
    \small
    \vspace{1mm}
    \caption{Channel configurations of different widening strategies}
    \begin{tabular}{c|c|c|c|c}
    \toprule
    Layers & W1-High & W1-Mid & W4-Low & W13-All \\
    \midrule
    Conv1 & 64 & 64 & 512 & 128 \\
    Conv2 & 64 & 64 & 512 & 128 \\
    \midrule
    Max Pooling & - & - & - & - \\
    \midrule
    Conv3 & 128 & 128 & 615 & 256 \\
    Conv4 & 128 & 128 & 615 & 256 \\
    \midrule
    Max Pooling & - & - & - & - \\
    \midrule
    Conv5 & 256 & 2990 & 256 & 405 \\
    Conv6 & 256 & 256 & 256 & 405 \\
    Conv7 & 256 & 256 & 256 & 405 \\
    \midrule
    Max Pooling & - & - & - & - \\
    \midrule
    Conv8 & 512 & 512 & 512 & 615 \\
    Conv9 & 512 & 512 & 512 & 615 \\
    Conv10 & 512 & 512 & 512 & 615 \\
    \midrule
    Max Pooling & - & - & - & - \\
    Conv11 & 1536 & 512 & 512 & 615 \\
    Conv12 & 512 & 512 & 512 & 615 \\
    Conv13 & 512 & 512 & 512 & 615 \\
    \midrule
    Max Pooling & - & - & - & - \\
    Soft-max & - & - & - & - \\
    \bottomrule
    \end{tabular}
    \label{tab:config}
\end{table}

\end{document}